\documentclass[runningheads]{llncs}
\usepackage[T1]{fontenc}
\usepackage{amsmath}
\usepackage{graphicx}
\usepackage{hyperref}

\usepackage{xcolor}
\usepackage{xspace}
\usepackage{newtxtext}
\usepackage{multirow}
\usepackage{color}

\usepackage{ragged2e}
\usepackage{tabularx}
\usepackage{makecell}
\usepackage{booktabs} 
\usepackage{cellspace} 
\usepackage{siunitx}

\newcommand{\tabArrayStretch}{1.0}
\newcommand{\tabVspaceBefore}{-.2cm} 
\newcommand{\tabVspaceAfter}{-.1cm} 
\newcommand{\figVspaceAfter}{0cm}
\newcommand{\secVspaceAfter}{-.1cm}
\newcommand{\parVspaceBefore}{-.1cm}
\newcommand{\titleVspaceBefore}{-.4cm}
\newcommand{\abstractVspaceBefore}{-.4cm}

\usepackage[numbers]{natbib}

\begin{document}

\title{Machine Learning and Statistical Insights into Hospital Stay Durations: The Italian EHR Case\vspace{\titleVspaceBefore}}
\titlerunning{ML and Statistical Insights into Hospital Stay Durations: The Italian EHR Case}
%
\author{Marina Andric\inst{1} \and Mauro Dragoni\inst{1}\orcidID{0000-0003-0380-6571}
}
%
\authorrunning{M. Andric, M. Dragoni}
%
\institute{Fondazione Bruno Kessler, Trento, Italy \\
\email{\{mandric,dragoni\}@fbk.eu}}

\maketitle  
\vspace{\abstractVspaceBefore}
\begin{abstract}



Length of hospital stay is a critical metric for assessing healthcare quality and optimizing hospital resource management. This study aims to identify factors influencing LoS within the Italian healthcare context, using a dataset of hospitalization records from over 60 healthcare facilities in the Piedmont region, spanning from 2020 to 2023. We explored a variety of features, including patient characteristics, comorbidities, admission details, and hospital-specific factors. Significant correlations were found between LoS and features such as age group, comorbidity score, admission type, and the month of admission.
Machine learning models, specifically CatBoost and Random Forest, were used to predict LoS. The highest R$^2$ score, 0.49, was achieved with CatBoost, demonstrating good predictive performance. 

\end{abstract}

\section{Introduction}
\label{sec:introduction}
\vspace{\secVspaceAfter}

Length of stay (LoS) is an important indicator of the use of medical services that is used to assess the efficiency of hospital management, patient quality of care, and functional performance~\cite{Gokhale2023}. Shorter LoS has been associated with decreased risks of infections, improvements in treatment outcomes, reduced medical costs, and increased hospital bed turnover~\cite{Bueno2010,Rotter2010}.

LoS varies even among patients with the same condition or undergoing the same type of intervention due to a complex interplay of individual characteristics, differences in medical practices across facilities~\cite{Baek2018}, and external influences like seasonal trends and public health crises. Understanding LoS requires accounting for these complexities. 
To explore these factors, this study leveraged electronic health records (EHR) from the TrustAlert research project,\footnote{\url{https://www.trustalert.it/}} an Italian initiative focused on developing early warning, monitoring, and forecasting tools for public health agencies. Spanning four years (2020-2023) and 66 healthcare facilities, the dataset offers a comprehensive view of both patient and hospital-related factors. 
This enabled a systematic analysis of factors influencing LoS, across time and institutions, with the goal of providing practical insights to help hospital administrators optimize inpatient care.

Our methodology (Section~\ref{sec:methodology}) began with selecting relevant EHR data (Section~\ref{sec:data}) and transforming it into a structured dataset of patient, admission, and hospital features (Section~\ref{sec:predictive-features}). These features formed the basis of two analyses: (1) a statistical analysis examining their relationship with LoS (Section~\ref{sec:statistical-analysis}) and (2) a predictive modeling approach using machine learning—Random Forest and CatBoost regression models—with different encoding techniques, including pretrained embeddings for diagnosis and procedure codes (Section~\ref{sec:predictive-modeling}).

Our results (Section~\ref{sec:results}) revealed significant statistical correlations between LoS and selected features (Section~\ref{sec:statistical-analysis-results}) and demonstrated strong predictive performance, with CatBoost consistently outperforming Random Forest and all features contributing to model accuracy (Section~\ref{sec:prediction-results}).
The study's conclusions (Section~\ref{sec:conclusions}) offer valuable insights for future LoS analysis and prediction.

\section{Methodology}
\label{sec:methodology}
\vspace{\secVspaceAfter}

\subsection{Data}
\label{sec:data}
\vspace{\secVspaceAfter}

\begin{figure}[t]
\centering
\includegraphics[width=\linewidth]{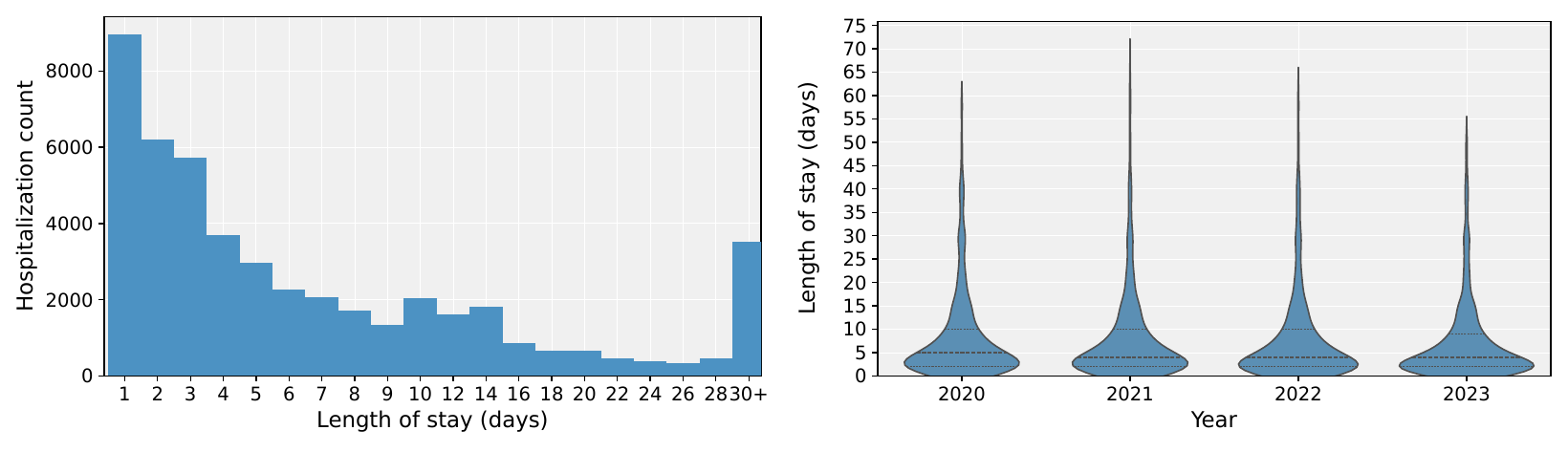}
\vspace{-.75cm}
\caption{Distribution of length of stay in days overall (left) and for specific years (right). In the violin plots, observations above the 98th percentile have been excluded for clarity, while black horizontal bars denote the 25th, 50th (median), and 75th percentiles.}
\label{fig:los-distribution}
\vspace{\figVspaceAfter}
\end{figure}

We analyzed inpatient hospitalization records from January 2020 to December 2023, extracted from EHRs of 66 healthcare facilities across 52 departments in Piedmont, Italy. Table~\ref{tab:data_description} summarizes dataset characteristics, while Table~\ref{tab:healthcare_facilities} details facility categories.

The dataset includes 59,684 hospitalizations ($\geq$ 1 day) involving 37,526 patients. Admission and discharge dates enabled LoS calculation, while de-identified patient IDs allowed hospitalization history reconstruction. Each record contained age group, hospitalization type, main diagnosis, main procedure, hospital, and department. 
Diagnoses and interventions were coded according to the 9th revision of the International Classification of Diseases (ICD-9), including the ICD-9 Procedure Coding System (PCS). 


\begin{table}[t]
    \centering
    \caption{Summary of data set characteristics (example data is artificial).}
    \label{tab:data_description}
     \vspace{\tabVspaceBefore}
    \footnotesize
    \renewcommand{\arraystretch}{\tabArrayStretch}
    \begin{tabularx}{\textwidth}{>{\raggedright\arraybackslash}p{3cm} *{2}{>{\raggedright\arraybackslash}X}}
        \toprule
        \textbf{Variable} & \textbf{Value} & \textbf{Example} \\
        \midrule
        Hashed patient ID & 37,526 unique patients & JL(\&LFURP(!2 \\
        Age group & 20 categories & 35-39\\ 
        Diagnosis code & 3,689 ICD-9 unique codes & 724.2 - Lumbago\\ 
        Procedure code & 1,645 ICD-9 unique codes & 88.93 - MRI of the Spinal Canal \\
        Facility & 66 unique values & Az. Ospedal. S. Croce E Carle \\
        Department & 52 unique values & Neurology \\
        Hospitalization type & Surgical, medical, ordinary & Medical \\
        Admission date & 01/01/2020 – 31/12/2023 & 22/07/2021 \\  
        Discharge date & 01/01/2020 – 31/12/2023 & 25/07/2021 \\       
        \bottomrule
    \end{tabularx}	
    \vspace{\tabVspaceAfter}
\end{table}

\begin{table}[t]
    \begin{minipage}{.455\linewidth}
        \caption{Facilities and \% samples by cat.}
        \label{tab:healthcare_facilities}
         \vspace{\tabVspaceBefore}
        \footnotesize
        \renewcommand{\arraystretch}{\tabArrayStretch}
        \setlength{\tabcolsep}{2pt}
        \begin{tabularx}{\linewidth}{@{}Xrr@{}}
            \toprule
            \textbf{Category} & \textbf{\#} & \textbf{\%} \\
            \midrule
            General/teaching hospitals                  & 13 & 16.4  \\  
            Specialized hospitals                           & 6  & 5.8   \\  
            Private clinics and rehab. centers      & 21 & 13.4  \\  
            Local/community hospitals                   & 19 & 62.6  \\  
            Geriatric/long-term care facilities         & 2  & 0.7   \\  
            Religious or charitable facilities      & 2  & 0.9   \\  
            Rehab. centers and research inst.  & 3  & 0.3   \\ 
            \textit{Total} & \textit{66} & \textit{59,684} \\
            \bottomrule
        \end{tabularx}
    \end{minipage}
    \hfill
    \begin{minipage}{.505\linewidth}
        \caption{Yearly LoS statistics.}
        \label{tab:los_statistics}
         \vspace{\tabVspaceBefore}
        \footnotesize
        \renewcommand{\arraystretch}{\tabArrayStretch}
        \setlength{\tabcolsep}{2pt}
        \begin{tabularx}{\linewidth}{@{}*{8}{>{\Centering}c}@{}}
            \toprule
            \textbf{Year} & \textbf{\#} & \textbf{\%} & \textbf{\hspace{-.2cm}Median\hspace{-.2cm}\,} & \textbf{Std} & \textbf{IQR} & \textbf{\hspace{-.075cm}Min\hspace{-.125cm}\,} & \textbf{Max} \\
            \midrule
            2020 & 10,147 & 17.0 & 5  & 20.80 & 2-11 & 1  & 362 \\
            2021 & 15,142 & 25.4 & 5  & 23.67 & 2-11 & 1  & 360 \\
            2022 & 16,674 & 27.9 & 4  & 22.74 & 2-10 & 1  & 357 \\
            2023 & 17,721 & 29.7 & 4  & 16.17 & 2-9  & 1  & 328 \\[3pt]
            \emph{all} & \hspace{-.1cm}59,684 & \hspace{-.1cm}100 & 4 & 20.94 & 2-10 & 1 & 362 \\
            \bottomrule
        \end{tabularx}
    \end{minipage}
    \vspace{\tabVspaceAfter}
\end{table}

Figure~\ref{fig:los-distribution} illustrates the distribution of the length of stay. The right panel shows the distribution of the length of stay separately for each year. 
In terms of hospitalization share, 2020 accounts for 17\% of sample, while 2021, 2022, and 2023 contribute approximately 25\%, 28\% and 30\% percent, respectively. 
Table~\ref{tab:los_statistics} shows key statistics for each year. Notably, for each following year, the plot shows broader section around the median, meaning higher concentration around the typical values. 
The years 2020 and 2021 exhibit larger interquartile ranges (IQRs) compared to 2022 and 2023, suggesting greater variability and unpredictability in the LoS during those earlier years.
These patterns likely reflect the impact of the COVID-19 pandemic. In 2020 and 2021, hospitals faced significant strain, leading to a reduction in non-COVID hospitalizations~\cite{Santi2021}. However, by 2022 and 2023, healthcare systems began to recover, resulting in more consistent hospitalization patterns.


\subsection{Features}
\label{sec:predictive-features}
\vspace{\secVspaceAfter}

In this study, we examined features broadly classified into three categories: patient-related, admission-related, and hospital-related (see Table~\ref{tab:variables-description}).
While previous research has explored these feature categories in LoS prediction (see~\cite{Gokhale2023,Stone2022} and references therein), existing studies often emphasize patient- and admission-related features, with less attention given to the role of hospital-related features and their temporal trends.

We transformed the raw hospitalization dataset into a structured feature set.
For each entry in the original dataset, a corresponding record was created in the feature dataset. In some cases, the data were used directly, while in others, we applied transformations that incorporated historical records, as detailed below. 
This feature set was then analyzed for its relationship with LoS using statistical and predictive methods.

\begin{table}[t]
    \centering
    \caption{Description of features used in the analysis.} 
    \label{tab:variables-description}
     \vspace{\tabVspaceBefore}
    \footnotesize
    \renewcommand{\arraystretch}{\tabArrayStretch}
    \begin{tabularx}{\textwidth}{>{\raggedright\arraybackslash}p{2.6cm} >{\raggedright\arraybackslash}X}
        \toprule
        \textbf{Variable} & \textbf{Description} \\ 
        \midrule
        Age group & Age category as a four-year interval (e.g., 0 for newborns, 1-4, up to 90+). \\
        \# of comorbidities & Count of unique comorbid conditions based on patient's past diagnoses. \\
        Comorbidity index & Elixhauser comorbidity index. \\
        Diagnosis & Main diagnosis associated with the hospitalization. \\
        Procedure & Main medical procedure performed during hospitalization. \\
        Admission type & Hospitalization type: day hospital (medical/surgical) or standard inpatient. \\
        Admission month & Month when the hospitalization occurred. \\
        Patient volume & Adjusted hospitalizations for a diagnosis in the past three months. \\
        Historical LoS & Average LoS for a diagnosis and hospital over the past three months. \\
        \bottomrule
    \end{tabularx}
    \vspace{\tabVspaceAfter}
\end{table}

\vspace{\parVspaceBefore}
\paragraph{Patient Features.}
The patient's age group was taken directly from the hospitalization record and was represented as a categorical variable based on four-year intervals, starting with \emph{0} (newborns), followed by \textit{1–4}, \textit{5–9}, and so on, up to \textit{90+}. 
We computed the number of comorbidities and the Elixhauser comorbidity index~\cite{Elixhauser1998}, which quantify a patient's overall health, and have been previously associated with LoS~\cite{Liu2019}.
These calculations were based on the patient's diagnosis history, which we constructed by extracting unique ICD-9 codes for prior diagnoses of a patient using admission dates to establish time ordering.
We leveraged Python's comorbidity (\texttt{comorbiPy}) library\footnote{\url{https://comorbidipy.readthedocs.io/en/latest/}} to calculate comorbidity measures.

\vspace{\parVspaceBefore}
\paragraph{Admission Features.} Diagnosis and procedure are clearly important features that influence care complexity and, consequently, the LoS. 
We explored different representations for diagnoses and procedures to determine their suitability for our analysis.
We used 100-dimensional diagnosis embeddings from Kane et al.~\cite{Kane2023} based on the BioGPT large language model~\cite{Luo2022}, and 300-dimensional procedure embeddings from Choi et al.~\cite{Choi2016}, trained with a neural language model. Since the diagnosis embeddings were originally designed for ICD-10 codes, we mapped ICD-9 codes to ICD-10 using the General Equivalence Mappings in Python’s \texttt{icdmappings} library\footnote{\url{https://pypi.org/project/icd-mappings/}}. We also tested one-hot and target encoding as alternative representations in preliminary experiments to identify the effective approach. Furthermore, we included month and type of admission to capture seasonal trends and variations in hospital practices.

\vspace{\parVspaceBefore}
\paragraph{Hospital Features.}
Previous studies suggest that higher-volume hospitals may have more experience, specialized expertise, and established care protocols~\cite{Kahn2006,Birkmeyer2002}. 
It is therefore plausible that hospital volume influences LoS.
We modeled the hospital patient volume feature as the number of patients admitted to the hospital with a specific diagnosis over the past three months, normalized by the total number of observation days.
To account for LoS variations across hospitals, we introduced a feature reflecting the average past hospitalization duration for each hospital, department, and diagnosis. 
For a given hospitalization record, we identified prior hospitalizations within the past three months that shared the same hospital, department, and diagnosis. 
Based on these historical records, we computed the observed mean LoS. To mitigate data scarcity, we also calculated a global mean LoS for related diagnoses using the \texttt{drgpy} library, which groups diagnoses into Diagnosis Related Groups (MS-DRG) based on hospital resource usage~\cite{DRGDesignDevelopment}. 
Finally, a smoothed mean was obtained by combining the observed and global means using a Bayesian approach with a smoothing parameter of five.

\subsection{Statistical Analysis}
\label{sec:statistical-analysis}
\vspace{\secVspaceAfter}
The statistical analysis explored features influencing LoS from multiple angles. 
Since diagnosis and procedure are known to be strong predictors of LoS~\cite{Stone2022}, they were excluded from our statistical analysis to focus on broader, less explored hospital-, admission-, and patient-related factors.

Using the feature set outlined in Section~\ref{sec:predictive-features}, the number of comorbidities and comorbidity index were discretized into categories~\cite{Austin2015}, while other categorical variables (age group, admission type, and admission month) were used directly. 
The relationship between each of these variables and LoS was then assessed using the Kruskal-Wallis test~\cite{Carter2014}, which is robust to non-normal distributions and suitable for comparing LoS across independent groups. 
Additionally, LoS variability within each category was summarized by the median, interquartile range, and standard deviation.
To examine the relationships between patient volume, historical LoS, and LoS, we employed mixed-effects linear models with LoS as the outcome variable. 
Separate models were trained using patient volume and historical LoS as independent variables, incorporating diagnosis as a random effect. 
To account for potential temporal variations, we explored interaction terms to assess how these relationships change over time.
Statistical significance was assessed at a 0.05 significance level to identify meaningful associations. The results are presented in Section~\ref{sec:statistical-analysis-results}.

\subsection{Predictive Modeling}
\label{sec:predictive-modeling}
\vspace{\secVspaceAfter}

\paragraph{Machine Learning Models.} We employed machine learning models to assess the predictive power of the considered features (see Table~\ref{tab:variables-description}). 
Due to likely complex, non-linear relationship between features and LoS, we framed the task as a non-linear regression problem and employed two robust tree-based models: Random Forest and CatBoost~\cite{Prokhorenkova2018}.
Both models efficiently handle large, diverse datasets and have been shown to outperform other regression models in predicting LoS~\cite{Jain2024}.
Interestingly, the models exhibited different sensitivities to how categorical features were represented. CatBoost performed best when diagnosis and procedure features were used as categorical variables, likely because of its built-in target encoding, which encodes each category based on the mean value of the target variable for that category. In contrast, Random Forest showed weaker performance when categorical features were used directly or one-hot encoded but performed best when represented as embeddings. 
In our results (Section~\ref{sec:prediction-results}), we present only the best-performing configurations: embeddings for Random Forest and categorical variables for CatBoost.

\vspace{\parVspaceBefore}
\paragraph{Validation Approach.} 
Given the potential seasonal trends exhibited by LoS, we split data into training, validation and test sets on a per-year basis to assess the models' capability to capture these variations. 
To assess the impact of additional historical data on model performance, we designed two \emph{split scenarios}. 
In the first, machine learning models were trained on hospitalization records from 2021, validated on 2022 records, and tested on 2023 records.
In the second, the training set was expanded to include records from both 2020 and 2021, while the validation and test sets remained the same (2022 and 2023, respectively).
This comparison enabled us to evaluate whether incorporating additional historical data from further in the past improves model performance.
For hyperparameter tuning, we conducted a grid search to evaluate various hyperparameter combinations and determine the best configuration for each model.

\vspace{\parVspaceBefore}
\paragraph{Evaluation Metrics.} To evaluate the quality of fit, we used adjusted R$^2$ and mean absolute error (MAE). We assessed these metrics for both the training and validation sets. Adjusted R$^2$ measures the proportion of variance explained by the model. Because it accounts for the number of predictors, it is more suitable than standard R$^2$ when comparing models with different feature sets~\cite{james2013introduction}. A consistently high adjusted R$^2$ value across the training and validation sets suggests that the model captures a significant portion of variability without overfitting. On the other side, MAE offers a straightforward measure of prediction accuracy by averaging the absolute differences between predicted and actual values. To evaluate the generalizability of the models on the test sets, in addition to MAE we used also root mean squared error (RMSE), which in contrast to MAE, penalizes large errors more heavily due to squaring~\cite{bishop2006pattern}.

\section{Results}
\label{sec:results}
\vspace{\secVspaceAfter}

\subsection{Statistical Analysis}
\label{sec:statistical-analysis-results}
\vspace{\secVspaceAfter}

\begin{table}[t!]
    \setlength{\tabcolsep}{5.5pt}
    \centering
    \caption{Patient statistics for different variables across all inpatient admissions. All analyzed variables showed a statistically significant relationship with LoS (p < 0.001)}
    \label{tab:patient-related-stats}
     \vspace{\tabVspaceBefore}
    \footnotesize
    \renewcommand{\arraystretch}{\tabArrayStretch}
    \setlength{\tabcolsep}{4pt}
    \begin{tabularx}{\textwidth}{X*{3}{c}@{\hspace{16pt}}*{5}{c}}
        \toprule
        \multirow{2}{*}{\textbf{Variable}} & \multirow{2}{*}{\textbf{Type}} & \multirow{2}{*}{\textbf{Number}} & \multirow{2}{*}{\textbf{\%}} & \multicolumn{5}{c}{\textbf{Length of stay (days)}} \\ \cmidrule{5-9}
        & & & & \textbf{Median} & \textbf{Std} & \textbf{IQR} & \textbf{Min} & \textbf{Max} \\
        \midrule
        \multirow{20}{*}{\parbox{2cm}{\raggedright Age group}}
        & 0 & 3,215 & 5.39  & 3 & 4.50  & 2-4 & 1  & 158 \\
        & 1-4 & 2,215 & 3.71  & 3 & 7.67  & 2-4 & 1  & 108 \\
        & 5-9 & 898  & 1.50  & 2 & 22.41 & 1-4 & 1  & 349 \\
        & 10-14 & 599 & 1.00  & 2 & 25.11 & 1-4 & 1  & 357 \\
        & 15-19 & 927 & 1.55  & 3 & 29.55 & 1-6 & 1  & 316 \\
        & 20-24 & 1,183 & 1.98  & 3 & 31.24 & 1-6 & 1  & 352 \\
        & 25-29 & 1,824 & 3.06  & 3 & 22.25 & 2-5 & 1  & 343 \\
        & 30-34 & 2,646 & 4.43  & 3 & 13.08 & 2-4 & 1  & 351 \\
        & 35-39 & 2,485 & 4.16  & 3 & 20.71 & 2-5 & 1  & 348 \\
        & 40-44 & 1,976 & 3.31  & 3 & 22.14 & 1-6 & 1  & 355 \\
        & 45-49 & 2,510 & 4.21  & 3 & 26.02 & 1-9 & 1  & 342 \\
        & 50-54 & 3,184 & 5.33  & 3 & 29.75 & 1-9 & 1  & 352 \\
        & 55-59 & 4,092 & 6.86  & 4 & 21.55 & 1-11 & 1 & 338 \\
        & 60-64 & 4,374 & 7.33  & 4 & 24.11 & 2-11 & 1 & 339 \\
        & 65-69 & 4,793 & 8.03  & 5 & 19.97 & 2-12 & 1 & 322 \\
        & 70-74 & 5,558 & 9.31  & 6 & 22.71 & 2-14 & 1 & 362 \\
        & 75-79 & 5,734 & 9.61  & 7 & 18.50 & 3-14 & 1 & 359 \\
        & 80-84 & 5,566 & 9.33  & 8 & 19.79 & 4-15 & 1 & 362 \\
        & 85-89 & 3,893 & 6.52  & 9 & 17.69 & 5-16 & 1 & 350 \\
        & 90+ & 2,012 & 3.37  & 9 & 10.41 & 5-15 & 1 & 118 \\
        \midrule
        \multirow{4}{*}{\parbox{2cm}{\raggedright Number of comorbidities}}
        & 0 & 53,533 & 89.69 & 4 & 19.50 & 2-9 & 1 & 362 \\
        & 1 & 5,578 & 9.35 & 8 & 30.88 & 3-17 & 1 & 338 \\
        & 2 & 552 & 0.92 & 8 & 18.23 & 4-17 & 1 & 198 \\
        & 3+ & 21 & 0.04 & 8 & 8.89 & 4-19 & 1 & 31 \\
        \midrule
        \multirow{4}{*}{\parbox{2cm}{\raggedright Elixhauser comorbidity index}}
        & \makecell{Low (,2]} & 54,542 & 91.38 & 4 & 19.47 & 2-10 & 1 & 362 \\
        & \makecell{Moderate [3,5]} & 629 & 1.05 & 8 & 14.79 & 3-15 & 1 & 191 \\
        & \makecell{High [6,10]} & 3,165 & 5.30 & 7 & 35.86 & 2-17 & 1 & 338 \\
        & \makecell{Very High [11,)} & 1,348 & 2.26 & 9 & 27.07 & 4-18 & 1 & 323 \\
        \bottomrule
    \end{tabularx}
    \vspace{\tabVspaceAfter}
\end{table}

\begin{table}[t]
    \centering
    \caption{Admission-related statistics across all inpatient admissions. All analyzed features showed a statistically significant relationship with LoS (p < 0.001)}
    \label{tab:admission-related-stats}
     \vspace{\tabVspaceBefore}
    \footnotesize
    \renewcommand{\arraystretch}{\tabArrayStretch} 
    \setlength{\tabcolsep}{4pt}
    \begin{tabularx}{\textwidth}{X*{3}{c}@{\hspace{16pt}}*{5}{c}}
        \toprule
        \multirow{2}{*}{\textbf{Variable}} & \multirow{2}{*}{\textbf{Type}} & \multirow{2}{*}{\textbf{Number}} & \multirow{2}{*}{\textbf{\%}} & \multicolumn{5}{c}{\textbf{Length of stay (days)}} \\ \cmidrule{5-9}
        & & & & \textbf{Median} & \textbf{Std} & \textbf{IQR} & \textbf{Min} & \textbf{Max} \\
        \midrule
        \multirow{11}{*}{\parbox{2cm}{\raggedright Month of admission}}
        & January & 4,641 & 7.78 & 5 & 44.62 & 2-13 & 1 & 362 \\
        & February & 4,617 & 7.74 & 4 & 26.07 & 2-10 & 1 & 323 \\
        & March & 5,175 & 8.67 & 4 & 23.04 & 2-11 & 1 & 294 \\
        & April & 4,381 & 7.34 & 5 & 20.60 & 2-12 & 1 & 267 \\
        & May & 5,076 & 8.50 & 5 & 18.93 & 2-11 & 1 & 232 \\
        & June & 5,079 & 8.51 & 4 & 17.41 & 2-10 & 1 & 198 \\
        & July & 5,329 & 8.93 & 4 & 18.77 & 2-11 & 1 & 180 \\
        & August & 4,509 & 7.55 & 5 & 16.08 & 2-11 & 1 & 146 \\
        & September & 5,270 & 8.83 & 4 & 13.35 & 2-10 & 1 & 112 \\
        & October & 5,900 & 9.89 & 4 & 12.24 & 2-10 & 1 & 90 \\
        & November & 5,600 & 9.38 & 4 & 9.03 & 2-10 & 1 & 58 \\
        & December & 4,107 & 6.88 & 3 & 4.79 & 2-7 & 1 & 30 \\
        \midrule
        \multirow{3}{*}{\parbox{2cm}{\raggedright Type of admission}}
        & Surgical & 8,090 & 13.55 & 1 & 11.82 & 1-1 & 1 & 253 \\
        & Medical & 3,777 & 6.33 & 7 & 58.92 & 2-35 & 1 & 362 \\
        & Ordinary & 47,817 & 80.12 & 5 & 14.34 & 3-11 & 1 & 359 \\
        \bottomrule
    \end{tabularx}
    \vspace{\tabVspaceAfter}
\end{table}

Tables~\ref{tab:patient-related-stats} and~\ref{tab:admission-related-stats} show the descriptive statistics for analyzed features.

The largest patient cohort, aged 70–84 years, made up about 28\% of all cases. The youngest group (0–14 years) had a narrow IQR (1–4 days), indicating short, predictable stays. Young adults (15–24 years) showed the highest LoS variability, with a standard deviation of 30 days, while the oldest patients (90+) had the most predictable stays, with a standard deviation of around 10 days.

Around 90\% of hospitalizations involved patients with no comorbidities, with a median LoS of 4 days. 
The presence of comorbidities doubled the median LoS to 8 days, with no significant change across 1, 2, or 3+ comorbidities. 
A higher Elixhauser index was associated with significantly longer and more variable stays.

Hospital admissions peaked in October–November and dropped sharply in December. January and February had the highest LoS variability, while December, November, and October showed the lowest variability.

Ordinary admissions, making up ~80\% of cases, had a median LoS of 5 days. Surgical cases had the shortest and most predictable stays (median: 1 day), while medical cases had the longest and most variable stays.

Age group, number of comorbidities, comorbidity index, admission type, and month all showed a statistically significant relationship with LoS (p < 0.001).

The mixed-effects model results indicated that in 2020, higher patient volume was associated with shorter LoS, but this relationship weakened or reversed in the following years, 2021, 2022, and 2023, with higher volume leading to longer LoS over time. Additionally, the influence of past hospitalization durations on current LoS has grown stronger in recent years. These trends were supported by statistically significant interaction terms at the 0.05 level.


\begin{figure}[h!]
    \centering
    \includegraphics[width=\linewidth]{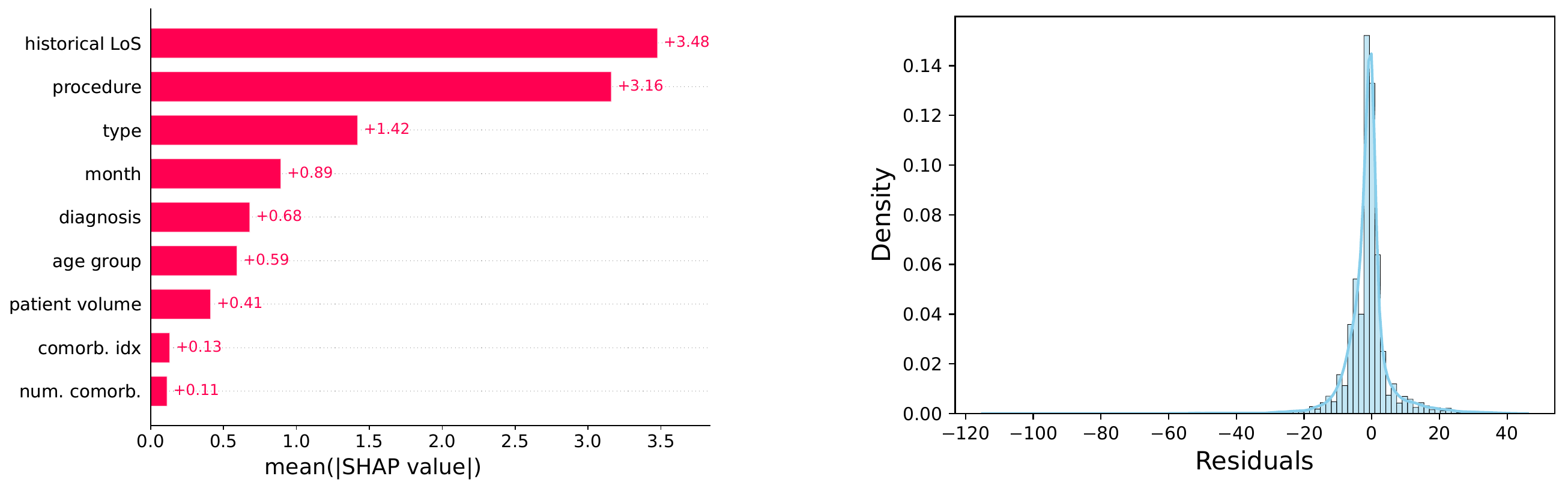}
    \vspace{-.5cm}
    \caption{SHAP feature importance bar plot (Left) and residuals distribution histogram (Right) for the CatBoost model trained on 2021 data with all features. Outliers below the 2nd and above the 98th percentile were removed from the histogram for clarity.}
    \label{fig:catboost-shap-learning}
    \vspace{\figVspaceAfter}
\end{figure}

\subsection{LoS Prediction Performance}
\label{sec:prediction-results}
\vspace{\secVspaceAfter}
We conducted experiments with six distinct feature sets to assess their impact on model performance. 
Starting with patient characteristics and diagnosis, we incrementally added procedure, admission type, month, patient volume, and historical LoS, evaluating models across two split scenarios. 
Results for Random Forest and CatBoost are shown in Table~\ref{tab:performance-results}.  

\setlength{\tabcolsep}{4.15pt} 
\begin{table}[t]
    \centering
    \caption{Performance of regression models on training, validation, and test datasets across two evaluation splits. Acronyms: RF – Random Forest, CB – CatBoost.} 
    \label{tab:performance-results}
     \vspace{\tabVspaceBefore}
    \footnotesize
    \renewcommand{\arraystretch}{\tabArrayStretch} 
    \setlength{\tabcolsep}{3pt}
    \begin{tabularx}{\textwidth}{ 
        @{}
        l@{\hspace{15pt}}>{\RaggedRight}X
        *{2}{@{\hspace{18pt}}rr@{\hspace{9pt}}rr@{\hspace{9pt}}rr}
        @{}
    }
        \toprule
        \multirow{3}{*}{\textbf{R\#}}
        & \multirow{3}{*}{\textbf{\hspace{-6pt}Model}}
        & \multicolumn{6}{c}{\textbf{\hspace{-18pt}Train: 2021, Val: 2022, Test: 2023}}
        & \multicolumn{6}{c}{\textbf{\hspace{-9pt}Train:\,2020-21, Val:\,2022, Test:\,2023}\hspace{-9pt}\,} \\
        \cmidrule(r{18pt}){3-8} \cmidrule{9-14} &
        & \multicolumn{2}{c}{\textbf{\hspace{-9pt}Train}}
        & \multicolumn{2}{c}{\textbf{\hspace{-9pt}Validation}}
        & \multicolumn{2}{c}{\textbf{\hspace{-15pt}Test}}
        & \multicolumn{2}{c}{\textbf{\hspace{-9pt}Train}}
        & \multicolumn{2}{c}{\textbf{\hspace{-9pt}Validation}}
        & \multicolumn{2}{c}{\textbf{Test}} \\
        \cmidrule(r{9pt}){3-4} \cmidrule(r{9pt}){5-6} \cmidrule(r{18pt}){7-8}
        \cmidrule(r{9pt}){9-10} \cmidrule(r{9pt}){11-12} \cmidrule{13-14} &
        & \textbf{\textsc{mae}} & \textbf{\textsc{r}$^2$}
        & \textbf{\textsc{mae}} & \textbf{\textsc{r}$^2$}
        & \textbf{\textsc{mae}} & \textbf{\textsc{rmse}}
        & \textbf{\textsc{mae}} & \textbf{\textsc{r}$^2$}
        & \textbf{\textsc{mae}} & \textbf{\textsc{r}$^2$}
        & \textbf{\textsc{mae}} & \textbf{\textsc{rmse}} \\
        \midrule
        \multicolumn{2}{c}{} & \multicolumn{12}{c}{patient feat. + diagnosis} \\[2.5pt]
        1 & RF
          & 8.40 & 0.29 & 8.62 & 0.21 & 7.71 & 14.74 
          & 8.26 & 0.28 & 8.50 & 0.22 & 7.62 & 14.71 
        \\
        2 & CB
          & 7.37 & 0.26 & 7.64 & 0.16 & 6.82 & 15.40 
          & 7.17 & 0.26 & 7.34 & 0.19 & 6.41 & 14.70 
        \\
        \midrule
        \multicolumn{2}{c}{} & \multicolumn{12}{c}{patient feat. + diagnosis + procedure} \\[2.5pt]
        3 & RF
          & 7.79 & 0.33 & 8.12 & 0.19 & 7.25 & 14.26 
          & 7.67 & 0.31 & 7.91 & 0.20 & 7.07 & 14.11 
        \\
        4 & CB
          & 6.79 & 0.34 & 7.02 & 0.23 & 6.28 & 14.59 
          & 6.78 & 0.33 & 6.72 & 0.27 & 5.89 & 13.92 
        \\
        \midrule
        \multicolumn{2}{c}{} & \multicolumn{12}{c}{patient feat. + diagnosis + procedure + type} \\[2.5pt]
        5 & RF
          & 7.77 & 0.35 & 8.06 & 0.22 & 7.24 & 14.12 
          & 7.63 & 0.34 & 7.85 & 0.24 & 7.07 & 13.95 
        \\
        6 & CB
          & 6.58 & 0.43 & 6.94 & 0.34 & 6.09 & 13.98 
          & 6.33 & 0.44 & 6.73 & 0.36 & 5.90 & 13.76 
        \\
        \midrule
        \multicolumn{2}{c}{} & \multicolumn{12}{c}{patient feat. + diagnosis + procedure + type + month} \\[2.5pt]
        7 & RF
          & 7.75 & 0.36 & 8.08 & 0.22 & 7.27 & 14.18 
          & 7.65 & 0.33 & 7.89 & 0.23 & 7.10 & 14.00 
        \\
        8 & CB
          & 6.40 & 0.48 & 6.90 & 0.38 & 6.17 & 14.06 
          & 6.17 & 0.51 & 6.68 & 0.42 & 5.86 & 13.34 
        \\
        \midrule
        \multicolumn{2}{c}{} & \multicolumn{12}{c}{patient feat. + diagnosis + procedure + type + month + patient vol.} \\[2.5pt]
        9 & RF
          & 7.77 & 0.35 & 8.07 & 0.22 & 7.26 & 14.13 
          & 7.65 & 0.32 & 7.88 & 0.23 & 7.07 & 13.97 
        \\
        10 & CB
          & 5.93 & 0.55 & 6.74 & 0.39 & 6.03 & 13.69 
          & 6.20 & 0.51 & 6.64 & 0.43 & 5.88 & 13.13 
        \\
        \midrule
        \multicolumn{2}{c}{} & \multicolumn{12}{c}{patient feat. + diagnosis + procedure + type + month + patient vol. + hist. LoS} \\[2.5pt]
        11 & RF
          & 7.58 & 0.39 & 7.83 & 0.26 & 6.99 & 13.58 
          & 7.52 & 0.36 & 7.71 & 0.27 & 6.88 & 13.49 
        \\
        12 & CB
          & 5.66 & 0.60 & 6.04 & 0.49 & 4.88 & 11.44 
          & 5.73 & 0.57 & 6.01 & 0.49 & 4.91 & 11.62 
        \\
        \bottomrule                
        
            
    \end{tabularx}
    \vspace{\tabVspaceAfter}
\end{table}

Overall, CatBoost consistently outperformed Random Forest across all metrics, feature combinations, and evaluation scenarios.
Further analysis of CatBoost revealed that adding the procedure feature alongside patient characteristics and diagnosis improved performance, as reflected by reduced validation and test MAE/RMSE and increased R² (rows 2 and 4).
Incorporating admission type significantly boosted validation R² from 0.23 to 0.34 (rows 4 and 6), while validation MAE remained stable. Test results supported this improvement, with RMSE decreasing from 14.59 to 13.89 in the first evaluation split and from 13.92 to 13.76 in the second.
Adding admission month had the most impact in the two-year split scenario, where validation R² rose from 0.36 to 0.42, and test RMSE dropped from 13.76 to 13.34 (rows 6 and 8). This suggests that a larger dataset helps the model capture seasonal trends more effectively.
The inclusion of patient volume led to further performance gains (rows 8 and 10).
Finally, historical LoS provided the greatest improvement (rows 10 and 12). However, training on both 2020 and 2021 data did not outperform using 2021 alone on the 2023 test set, likely due to shifts in patterns from 2020 making the model less representative (see Section~\ref{sec:statistical-analysis-results}).

Figure~\ref{fig:catboost-shap-learning} illustrates the SHAP feature importance bar plot (Left) and residuals distribution histogram (Right) for the best-performing CatBoost model (trained on 2021 data with all features). 
Historical length of stay and procedure were the most influential features, which is consistent with expectations, as both are strong predictors of a patient’s length of stay. On the other hand, the number of comorbidities and the Elixhauser comorbidity index show minimal impact, which is expected due to the highly skewed distribution of these values in our dataset.
The residuals exhibit an approximately normal distribution, indicating that the model is well-fitted with random, unbiased errors. The distribution is centered around zero, which suggests there is no significant bias in the predictions.



\section{Conclusions}
\label{sec:conclusions}
\vspace{\secVspaceAfter}

Understanding hospital length of stay is important for optimizing resource management and improving patient care. By identifying key factors influencing LoS, hospitals can better predict and manage inpatient days.

This study examined patient, admission, and hospital-related features for LoS prediction, with CatBoost consistently outperforming Random Forest and improving as more features were added. Historical LoS was the most influential feature.
Notably, in 2020, higher patient volume correlated with shorter LoS, but from 2021 onward, the trend reversed, suggesting that increasing patient loads may have strained hospital capacity over time.
With all features included, CatBoost achieved the highest validation R$^2$ of 0.49, indicating moderate predictive power. These findings align with prior research~\cite{Baek2018} on hospital processes and suggest that integrating additional predictive features could further enhance LoS modeling.

Additionally, target encoding for diagnosis and procedures outperformed previously published embeddings for LoS prediction with CatBoost. Future work could explore task-specific LLM embeddings to improve representation learning.

\bibliographystyle{splncs04}
\bibliography{bibliography}

\end{document}